\documentclass[letterpaper, 10 pt, conference]{ieeeconf}  

\IEEEoverridecommandlockouts                              

\usepackage{graphics} 
\usepackage{epsfig} 
\usepackage{mathptmx} 
\usepackage{times} 
\usepackage{amsmath} 
\usepackage{amssymb}  
\usepackage{graphicx}    
\usepackage{subcaption}

\usepackage{times} 
\usepackage{color,xcolor}
\usepackage{url}
\usepackage{booktabs}
\usepackage{float}
\usepackage{threeparttable}
\usepackage{multirow}
\usepackage{multicol}
\usepackage[final]{pdfpages}

\title{\LARGE \bf NF-SLAM: Effective, Normalizing Flow-supported Neural Field representations for object-level visual SLAM in automotive applications
}
\author{Li Cui$^{1}$, Yang Ding$^{1}$, Richard Hartley$^{2}$, Zirui Xie$^{1}$, Laurent Kneip$^{3,1}$ and Zhenghua Yu$^{1}$%
\thanks{$^{1}$Motovis Intelligent Technologies, $^{2}$Australian National University, $^{3}$ShanghaiTech University}}

\begin{document}
\maketitle
\thispagestyle{empty}
\pagestyle{empty}

\begin{abstract}
We propose a novel, vision-only object-level SLAM framework for automotive applications representing 3D shapes by implicit signed distance functions. Our key innovation consists of augmenting the standard neural representation by a normalizing flow network. As a result, achieving strong representation power on the specific class of road vehicles is made possible by compact networks with only 16-dimensional latent codes. Furthermore, the newly proposed architecture exhibits a significant performance improvement in the presence of only sparse and noisy data, which is demonstrated through comparative experiments on synthetic data. The module is embedded into the back-end of a stereo-vision based framework for joint, incremental shape optimization. The loss function is given by a combination of a sparse 3D point-based SDF loss, a sparse rendering loss, and a semantic mask-based silhouette-consistency term. We furthermore leverage semantic information to determine keypoint extraction density in the front-end. Finally, experimental results on real-world data reveal accurate and reliable performance comparable to alternative frameworks that make use of direct depth readings. The proposed method performs well with only sparse 3D points obtained from bundle adjustment, and eventually continues to deliver stable results even under exclusive use of the mask-consistency term.
\end{abstract}

\section{INTRODUCTION}

Beyond merely estimating ego-motion and environment geometry, we require Simultaneous Localization And Mapping (SLAM) algorithms for automotive applications to generate an object-level understanding of the scene and estimate the dynamics and shape of surrounding vehicles. In line with recent works such as DSP-SLAM~\cite{wang2021dsp}, we propose a novel vision-only object-level SLAM framework able to estimate the exact pose and shape of nearby vehicles.

The dominating shape representation embedded into the back-end optimization of object-level SLAM frameworks is currently given by implicit neural shape generators such as DeepSDF~\cite{park2019deepsdf}. The network consists of a Multi-Layer Perceptron (MLP) conditioned by an optimizable latent code to predict the distance to the surface for an input sampling point. The latter may be chosen continuously in space, thereby adding a lot of flexibility to the sampling scheme and the way in which such representations can be embedded into the back-propagating optimization thread of both vision-based frameworks such as DSP-SLAM~\cite{wang2021dsp}, or lidar-based frameworks such as TwistSLAM++~\cite{gonzalez2022twistslam++}. However, as pointed out by Duggal et al.~\cite{duggal2022mending}, the raw DeepSDF representation faces difficulties in the presence of sparse or noisy visual measurements, and also returns strongly varying optimization results for only small variations of the initial latent code. It comes as no surprise that in DSP-SLAM~\cite{wang2021dsp}---a framework that can be used both with and without lidar measurements--- vision-only results perform vastly inferior to their counterpart that makes use of direct and denser depth readings.

In light of these difficulties, we propose NF-SLAM, a novel object-level stereo visual SLAM framework that incorporates the following innovations:
\begin{itemize}
  \item A sparse front-end feature detector with an adaptively tuned extraction density making use of semantic segmentation to identify regions of interest (i.e., vehicle detections). While the output remains sparse, this step enriches the object-related 3D point cloud that can be obtained from bundle adjustment.
  \item A novel generative shape representation that augments DeepSDF by a normalizing flow network, thereby stabilizing and improving results obtained from sparse visual measurements. The representation is readily embedded into optimization and makes use of only 16-dimensional latent codes as opposed to the 64-dimensional codes used in DSP-SLAM~\cite{wang2021dsp}.
  \item A complete object-level SLAM framework that successfully marries these novel front and back-end modules with an incremental optimization objective. The latter makes use of an SDF loss, a sparse rendering loss, as well as a silhouette consistency term.
\end{itemize}

Our experimental analysis is divided into two parts. First, we test our normalizing flow-augmented neural implicit shape representation on both synthetic complete and partial depth readings, and demonstrate superior performance with respect to DeepSDF~\cite{park2019deepsdf}. Second, we test the complete framework on the KITTI~\cite{geiger2012we} benchmark. We set a new state-of-the-art over DSP-SLAM~\cite{wang2021dsp} in vision-only mode, and furthermore maintain competitive performance even when lidar information is added. We furthermore demonstrate reasonably good performance while making use of mask constraints, only. We believe that our proposition is of high interest in the ADAS community, which currently gains traction and tendentially restrains itself to the close-to-market alternative of vision-only exteroceptive perception.

\section{Related Work}
\noindent\textbf{Object-level SLAM:} Dame et al.~\cite{dame13} introduce one of the first visual SLAM solutions operating at the level of objects using DCTs of SDFs as a low-dimensional shape representation. Later on, SLAM++~\cite{salas2014} highlights the multi-view pose constraints emanating from object pose estimation and proposes object-level pose graph optimization while relying on a database of shape priors. Zhu et al.~\cite{zhu17} again perform shape optimization within a photometric bundle adjustment objective, and are the first to use a deep neural network as an internal representation (notably for point clouds). Deep-SLAM++~\cite{Hu_2019_ICCV} also employs a shape completion network inside SLAM, however performs refinement by discrete selection from multiple forward predictions rather than back-propagating optimization. Node-SLAM~\cite{sucar2020} is the first object-level SLAM framework making use of DeepSDF~\cite{park2019deepsdf} within optimization. The latter is also embedded into DSP-SLAM~\cite{wang2021dsp}. Liu et al.~\cite{liu2023mv} also make use of implicit neural shape representations, however focus on the prediction of latent codes from multi-sweep lidar scans rather than optimization.  More recently, new representations for uncertainty-aware shape representation and optimization were introduced by Liao and Waslander~\cite{liao2023uncertainty,liao2024multi}, though with a focus on more general shape reconstruction from RGBD rather than vision-only automotive applications. Finally, Han et al.~\cite{RO-MAP} propose the use of neural radiance fields for object-level estimation, and Kong et al.~\cite{kong23} propose to train an MLP for implicit scene representation during SLAM, thereby not imposing priors but merely regularization constraints. DSP-SLAM~\cite{wang2021dsp} remains the most related to the present work, and serves as our reference baseline. Note however that the framework optionally uses lidar points, and we compare against both its depth-aware and vision-only variants.

Although the focus of the present work lies on object reconstruction rather than dynamic vehicle perception, it is worth acknowledging the existence of solutions attempting the estimation of dynamic vehicles in the scene by employing trajectory priors~\cite{li2018,Hu_2019_ICCV,bescos21,gonzalez2022twistslam++}. TwistSLAM++~\cite{gonzalez2022twistslam++} in particular again uses DeepSDF to estimate object shapes from lidar points.

\noindent\textbf{Object shape representations:} Although many representations for image or point cloud-based shape prediction have been presented in the literature \cite{fan17,liao18,groueix18} (e.g. voxel-based, point-cloud based, mesh-based), the biggest advance in terms of differential shape generators that can be embedded as a regularizer into an object-level SLAM framework has been brought along by the introduction of implicit neural field representations. The latter have focussed on scene-level field representations for occupancy~\cite{mescheder19} or radiance~\cite{mildenhall20}, and finally---through the introduction of DeepSDF~\cite{park2019deepsdf}---individual object shapes. DeepSDF has been the subject of many follow-up contributions aiming at improving the accuracy and compactness of the representation. 

In a direct follow-up to DeepSDF~\cite{park2019deepsdf}, Chabra et al.~\cite{chabra20} employ the representation at scene-level by parametrizing and interpolating SDFs in near-surface voxels, only. 

In an aim to reduce dissimilarities between generated shapes and a shape collection, Zheng et al.~\cite{zheng2022} add an end-to-end trainable GAN network to classify shapes as real or fake from both local and global perspectives. In an aim to improve performance under noisy and sparse measurements, Duggal et al.~\cite{duggal2022mending} similarly propose the addition of an encoder for robust latent code initialization and a discriminator to induce a learned prior over the predicted SDF. While the addition of a discriminator is helpful, the presently proposed addition of a normalizing flow network forms a more natural architecture to be embedded into back-propagating optimization.

Recently, the introduction of transformers and diffusion models has lead to another wave of powerful differential shape generation networks ~\cite{hui2022,muller2022diffrf}, however moving away from the potentially compact and highly flexible implicit function representations. In an effort to reduce complexity, the community has proposed the tri-planes abstraction
~\cite{shue20233d,wang2023}. More recently, Yariv et al. propose Mosaic-SDF~\cite{yariv2023mosaicsdf}, which---in analogy to Deep Local Shapes~\cite{yariv2023mosaicsdf}---applies the transformer based diffusion model only in local sub-volumes near the surface. While this reduces the size of each tensorial sub-volume to 7$\times$7$\times$7, diffusion-based architectures remain computationally demanding as optimizing a single shape still takes more than 2 minutes on an A100 GPU. As a result, diffusion models are not yet amenable to real-time object-level SLAM.

\noindent\textbf{Normalizing flow:} 
The present work advocates the use of compact, application-specific DeepSDF deviates elegantly augmented by a normalizing flow network to generate reasonable shapes from sparse and noisy visual measurements, and improve convergence behavior of the optimization. Normalizing flow networks have already been promoted in point-cloud oriented representations such as PointFlow~\cite{yang2019pointflow}, DPF-Flow~\cite{klokov2020discrete}, SoftFlow~\cite{kim2020softflow}, and Go with the flows~\cite{postels2021go}. Here we make use of iterative \textit{Gaussianization Flow}~\cite{meng2020gaussianization} for latent shape code distribution learning. It uses a repeated composition of trainable kernel layers and orthogonal transformations, can transform any continuous random vector into a Gaussian one, and has demonstrated competitive performance versus several other representations (Real-NVP~\cite{dinh2016density}, Glow~\cite{kingma2018glow}, and FFJORD~\cite{grathwohl2018ffjord}).

\section{Introduction to {NF-SLAM}}
\begin{figure*}[htb]
\centering
\vspace{5pt}
\includegraphics[width=0.85\linewidth]{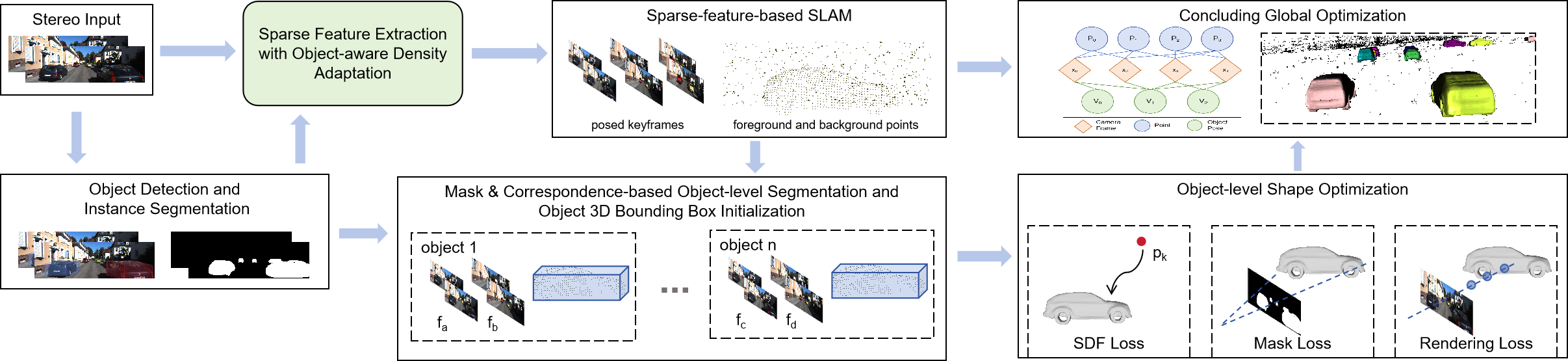}
\caption{Overview of our proposed framework for object-level visual SLAM in automotive applications.}
\label{fig: pipeline}
\vspace{-10pt}
\end{figure*}
As for many SLAM frameworks proposed over the past years, both DSP-SLAM and the presently proposed NF-SLAM make use of ORB-SLAM2~\cite{murORB2} as a sparse feature-based foundation for efficient tracking and mapping. DSP-SLAM furthermore incorporates lidar depth readings as an optional input, which turns out to be crucial in order to obtain sufficiently accurate object shape estimations. Our newly proposed framework uses only stereo images. However, by averaging temporal information and making use of a normalizing flow-supported implicit representation, we manage to achieve reliable results even in the absence of direct depth readings. In the following, we will go through the notations used in this paper, a complete system overview, as well as a summary of both the front-end and back-end modules.

\subsection{Notations and conventions}
The input of the proposed system is given by stereo images. We denote the left and right image of the $i$-th stereo pair $f_{li}$ and $f_{ri}$. In order to get semantic information, we utilize an off-the-shelf 2D detector to identify 2D bounding boxes $B_{ij}$ and instance masks $M_{ij}$ for each object in each image. All subsequent operations are conducted based on the left image as the reference. $T^c_o$ refers to the 7-DoF pose of the object. Pixels in the latter are denoted by vectors $u$. In order to represent shapes, our implicit distance function $F_{\theta}$ makes use of a latent shape code $\mathbf{z} \in \mathbb{R}^{16}$. Let $G_{\phi}$ furthermore be the normalizing flow network pre-pended to the MLP. Its input is given by the variable $\mathbf{w} \in \mathbb{R}^{16}$, which ideally obeys a normal distribution. $\theta$ and $\phi$ represent the weights of these two networks, respectively.
\vspace{-0.22cm}
\subsection{Framework overview}
An overview of the complete framework is illustrated in Figure~\ref{fig: pipeline}. Each newly incoming stereo frame is first subjected to front-end modules for detecting 2D bounding boxes of objects and their respective instance-level semantic masks~\cite{he2017mask} as well as sparse ORB keypoints. The latter are then forwarded to ORB-SLAM2~\cite{murORB2} for sparse, stereo visual SLAM. As a result of this module, we will obtain a sparse 3D point cloud and stereo frames all registered within a global reference frame. The segmentation module then checks in which frame and notably mask each of the optimized 3D points is visible, and forms new, instance-level sub-graphs containing adjacent stereo frames that observe the same object and its corresponding subset of 3D points. The extraction of adjacent temporal observations and object-level multi-view graphs is crucial for reliability in the vision-only case as reliance on a single stereo frame may easily lead to suboptimal results.

Once the subgraphs are defined, a combination of 3D point information as well as 2D bounding box coordinates is used in order to initialize a 3D bounding box around each object. Next, we formulate a shape optimization objective based on our normalizing flow-supported implicit neural representation for each individual object. Finally, the reconstructed objects are incorporated into a joint factor graph for global bundle adjustment, aiming to achieve a globally consistent map. Note that, while Figure~\ref{fig: pipeline} suggests sequential batch processing of all modules, the modules are indeed all operating incrementally and adding new observations and constraints as new stereo images are being processed.

For the sake of visualization purposes, a fast choice with online capability is given by the marching cubes algorithm.

\subsection{Details on front-end modules}

The standard ORB feature extractor from ORB-SLAM2 aims at an as-homogeneous-as-possible feature distribution, which may be benefitial for the robustness and accuracy of the ego-motion estimation. However, given that object observations often make up for only a small part of the image, it naturally limits the number of sparse keypoints that we extract on object surfaces, thereby limiting the potential given by some of the employed loss functions. We therefore implement a modified feature extractor that revisits 2D Regions Of Interest (ROIs) corresponding to object bounding boxes, and significantly enriches the feature extraction in those regions by applying a vastly reduced threshold. As a result, we naturally obtain more correspondences and optimized 3D points on object surfaces. This in turn provides the shape optimizer with an increased number of constraints, and as a result leads to better overall shape estimation results. Note that this operation is performed in both the left and the right image, which also explains the need to extract semantic information in both views.

A second module of interest in the front-end is given by the object pose initialization module. We make use of the 3D bounding box initialization method proposed by Li et al.~\cite{li2018}, which is taylored to stereo cameras. However, the method depends on a view-point classifier, which we replace by a discretization of the orientation coming out of a principle component analysis of the object's 3D point cloud.

\subsection{Details on back-end modules}

The back-end module consists of two sub modules. The first one takes sets of adjacent frames observing a single object along with the relative object frame definition, the 3D object points, corresponding image keypoints, the masks, and the bounding boxes, and feeds them to a per-object shape optimization module. The objective of the shape optimization is composed of up to three losses, meaning an SDF loss (i.e. sparse 3D point-to-surface distances), a silhouette inconsistency loss (i.e. discrepancies between rendering-based occupancies and measured occupancies), and a rendering loss (i.e. differences between rendered depths and optimized depths). Note that all three losses are formulated as a function of our normalizing-flow supported, implicit neural shape representation. The shape optimization method forms the core of our contribution, and more details on representation and optimization are provided in Section~\ref{chapter: shape optimizer}.

The concluding stage is given by global optimization of a joint factor graph that includes point features, camera poses, and object poses. The original graph optimization module from ORB-SLAM2 already performs bundle adjustment and loop closure to ensure a globally consistent map. As new objects are being detected, they are incorporated as further nodes in a joint factor graph, and the edges are formed by the corresponding relative object pose estimates. The main target of this extension is to improve the accuracy of the relative location of each object reference frame.

\section{Vehcile shape optimization with normalizing-flow supported implicit fields}
\label{chapter: shape optimizer}

Our proposed vehicle shape optimization makes use of an implicit neural shape generator. The present section introduces the architecture of the network, the constraints employed during optimization, and the concluding optimization itself.

\subsection{Network Architecture}
Our shape optimizer internally incorporates two types of networks: normalizing flow and DeepSDF. It aims to optimize latent vectors from a standard Gaussian distribution into a high-dimensional representation that best aligns with the current point cloud shape. For each queried point, it computes its signed distance function value.

\begin{figure}[t!]
    \vspace{5pt}
    \centering
    \includegraphics[scale=0.38]{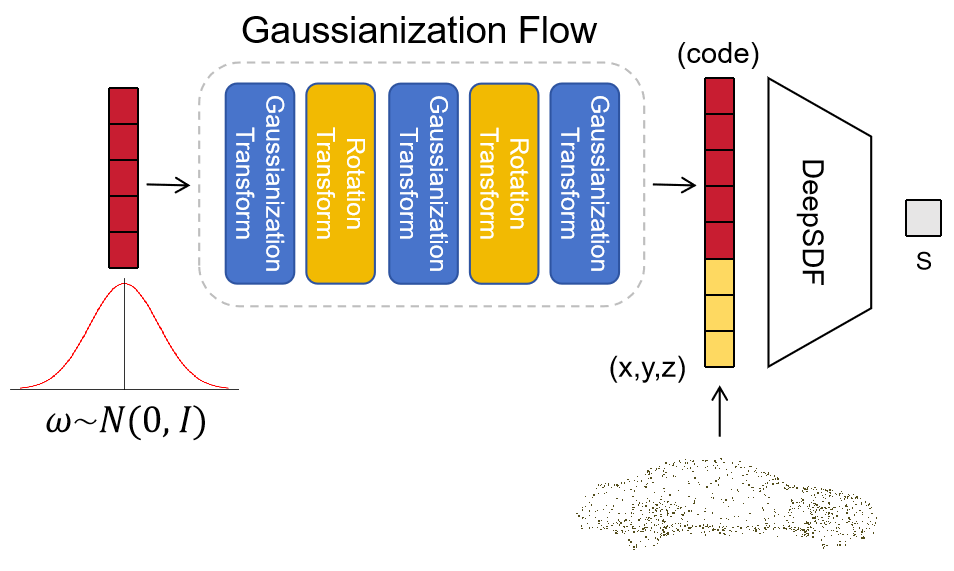}
    \caption{Implicit neural shape representation used in the present work. The architecture is composed of a DeepSDF decoder preceded by a normalizing flow network. Given an input latent code $\mathbf{w}$ and a sampling point $\{x,y,z\}$, the network generates the 3D Euclidean distance between the sampled point and the object surface.}
    \label{fig:testing}
    \vspace{-5pt}
\end{figure}

The normalizing flow module $G_{\phi}$ is designed to generate shape latent vectors from variables that adhere to a standard Gaussian distribution. Drawing inspiration from the work of Meng et al. \cite{meng2020gaussianization}, we use Gaussianization flow as our normalizing flow embedding. Gaussianization flow is a trainable extension of Rotation-Based Iterative Gaussianization. It is constructed by stacking trainable kernel layers and orthogonal matrix layers alternatively, as Fig.\ref{fig:testing} shows. 

We employ DeepSDF as our shape decoder $F_{\theta}$, which can estimate the SDF value for each point from a latent vector $\mathbf{z}\in\mathbb{R}^{16}$. The normalized code $\mathbf{w}$ hence also has only 16 dimensions. Our DeepSDF derivate has only 8 fully connected layers, and the normalizing flow module employs three Gaussianization transform layers with interleaving rotation transform layers. The network is trained in two stages. We first train the DeepSDF network, and subsequently train the normalizing flow network.

\subsection{Optimization Constraints}
During shape reconstruction, we fix the network weights $\theta, \phi$. For each instance, assuming $\mathbf{z}$ represents the latent shape code and $p_{\mathbf{z}}$ its data distribution, the normalizing flow module executes a bijective, differentiable transformation of $\mathbf{z}$ into $\mathbf{w} = G_{\phi}^{-1}(\mathbf{z})$. We first initialize $\mathbf{w}$ with zero-value vector from scratch. Subsequently, the latent code $\mathbf{z}$ can be generated using the normalizing flow network. Next, each test point is concatenated with $\mathbf{z}$, and jointly fed into DeepSDF. The trained DeepSDF is capable of predicting the SDF value for each test point $\mathbf{p}$, i.e.
\begin{equation}
\begin{aligned} 
s &=   F_{\theta}(T_c^o \mathbf{p},G_{\phi}(\mathbf{w})) 
\end{aligned}
\label{eq: sdf compuatation}
\end{equation}
Under the supervision of certain loss functions, we then try to find an optimal $\hat{\mathbf{w}}$ and $\hat{T_{o}^{c}}$. With the optimal shape code in hand, we can apply the marching cubes algorithm to get a full 3D mesh.

We utilize the following constraints for shape optimization.

\subsubsection{SDF surface constraint}
Each 3D point $\mathbf{p}_k$ corresponding to a detected sparse keypoint can be transformed into its corresponding canonical object reference frame using $P^o_k = T_c^o \mathbf{p}_k$. In an ideal scenario, the SDF value of points on the object surface is expected to be 0, so we can formulate the surface loss by
\begin{equation}
\begin{aligned} 
L_{surf} &=  \frac{1}{N} \sum_{k=1}^{N} \Vert F_{\theta}(T_c^o \mathbf{p}_{k},G_{\phi}(\mathbf{w}))  \Vert^2.
\end{aligned}
\label{eq:surface loss}
\end{equation}

\subsubsection{Foreground mask constraint}
\label{chap: Foreground mask constraint}
Inspired by work of Tulsiani et al.~\cite{tulsiani2017multi}, we formulate a differentiable loss function termed as 'view consistency', which quantifies the disparity between the reconstructed 3D shape and its corresponding observation in the image.
During the ray tracing process, we use $e_i$ to describe the probability that the $i^{th}$ sampling point is empty, where 1 means being empty and vice visa. According to Equation~\ref{eq:empty prob}, it can be computed from the SDF value $s_i$ by
\begin{equation}
e_i = \begin{cases}
0& \text{$s_i$ \textless -$\sigma$ }\\
\frac{1}{2} + \frac{s_i}{2 \sigma}& \text{$\lvert s_i \rvert$ $\leq \sigma$ }\\
1& \text{$s_i$ \textgreater $\sigma$}
\end{cases}.
\label{eq:empty prob}
\end{equation}
Suppose the ray $r$ passes through $N_r$ sampling points. The ray may hit one of the $N_r$ sampling points or just escape. We use the variable $t_r$ to represent the sampling point at which the ray probabilistically terminates, where $t_r = N_r + 1$ means the ray doesn't hit the surface
\begin{equation}
p(t_r = i) =
    \begin{cases}
    (1- e_i) \prod \limits_{j=1}^{i-1} e_j & \text{ $i\leq{N_r}$}\\
    \prod \limits_{j=1}^{N_r} e_j & \text{$i=N_r+1$}
    \end{cases}
\label{eq: event prob}
\end{equation}
To combine the known information from the object mask for each ray, we use $m_r \in \left\{0,1 \right\}$, where $m_r = 0$ means the ray $r$ intersects corresponding to a pixel within the object mask and thus is supposed to hit the surface, and $m_r = 1$ means the ray doesn't hit the target object. We can finally assign a cost $\psi_r^{mask}(i)$ to the event $(t_r = i)$, thereby punishing predictions  inconsistent with observations. We have
\begin{equation}
\psi_r^{mask}(i) = 
    \begin{cases}
    m_r & \text{$i\leq N_r$ }\\
    1-m_r & \text{$i=N_r+1$}
    \end{cases}.
\end{equation}
Finally, our mask constraint is given by,
\begin{equation}
    \begin{aligned}
    L_{mask} &=  \frac{1}{\vert \Omega_{MB} \vert}\sum_{r \in \Omega_{MB}}\sum_{i=1}^{N_r+1} p(t_r=i) \psi_r^{mask}(i) 
    \end{aligned},
    \label{eq:mask loss}
\end{equation}
where $\Omega_{MB} = \Omega_{M} \cup \Omega_{B}$ means the collection of pixels located within the mask region and those outside the Mask region but inside the 2D bounding box.

\subsubsection{Depth constraint}
After estimating the depth $d_u$ of points through stereo triangulation or bundle adjustment, we can formulate an additional loss that considers the rendered depth at such points. Note that such rendering loss is not to be confused with the surface loss, as the rendering loss implicitly forces observed depth points to align with the unoccluded part of the surface, while the surface loss simply chooses the distance to the nearest point anywhere on the surface. During ray tracing along each ray, each sampled depth is $d_i= d_{min} + (i-1) \Delta d$, and the sampling interval step size is typically denoted by $\Delta d = (d_{max} - d_{min})/(N_r - 1)$. 
By combining Equation~\ref{eq: event prob},  we can compute the rendered depth for each pixel. The rendering depth loss is finally given by
\begin{equation}
    \begin{aligned}
    L_{depth} &= \frac{1}{\vert \Omega_{SB} \vert}\sum_{u \in \Omega_{SB}}(d_u - \sum_{i=1}^{N_r+1}   p_i d_i )^2
    \end{aligned},
    \label{eq:depth loss}
\end{equation}
where $\Omega_{SB} = \Omega_{S} \cup \Omega_{B}$ consists of points with depth lying on the object surface, as well as points not on the surface but still within the 2D bounding box.

\subsection{Optimization Process}
Due to the inaccuracies in single-frame image observations, we devise a joint multi-frame optimization strategy to reconstruct the shape. 
The points utilized in the surface constraints are derived from the fusion of 3D points across $Q$ frames. We compute the mask loss for each frame, and average them to obtain the final mask term for the object. A similar strategy is applied for the depth term. The last term is the regularization term.
\begin{equation}
\begin{aligned} 
L &= \lambda_1L_{surf} +  \sum_{i=1}^{Q} \big(\lambda_2 L_{mask} +  \lambda_3 L_{depth} \big)
 + \lambda_4 \Vert \mathbf{w} \Vert_2^2
\end{aligned}
\label{eq:whole loss}
\end{equation}
Given the quadratic nature of all terms, we apply a Gauss-Newton optimization strategy employing analytical Jacobians.

\section{Experiments}

Our experiments split up into two parts. We first test our newly proposed implicit neural shape representation and compare it against similarly dimensioned variants of Deep SDF. Second, we test the entire object-level SLAM framework on KITTI sequences and compare results against DSP-SLAM.

\subsection{Shape generator evaluation on synthetic datasets}
To assess whether the reconstructed shape closely matches the original point cloud, we conduct experiments on the synthesized dataset ShapeNet~\cite{chang2015shapenet}. We retrained the DeepSDF network with a 16 dimensional latent code and compare our new architecture against this variant to solely assess the impact of adding the normalizing flow network. Additionally, for both networks, we evaluate the impact of different optimization methods during inference, including Gauss-Newton and the Pytorch Adam optimizer.
\begin{figure*}[htb]
\centering
\subfloat[GT obj]{
\begin{minipage}[b]{0.15\linewidth}
\includegraphics[width=1\linewidth]{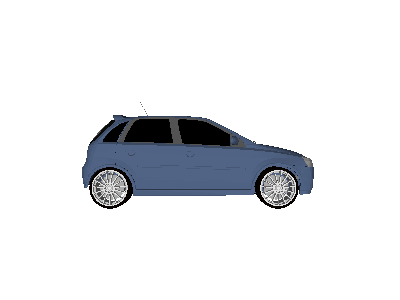}\vspace{-30pt}
\includegraphics[width=1\linewidth]{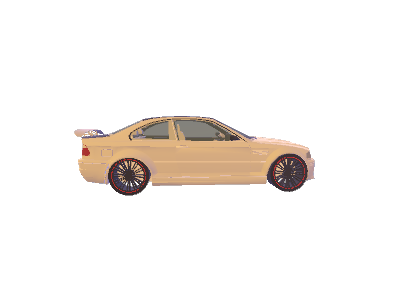}\vspace{-30pt}
\includegraphics[width=1\linewidth]{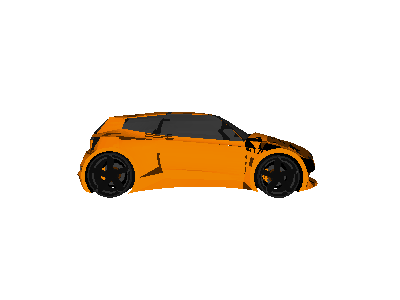}\vspace{-30pt}
\includegraphics[width=1\linewidth]{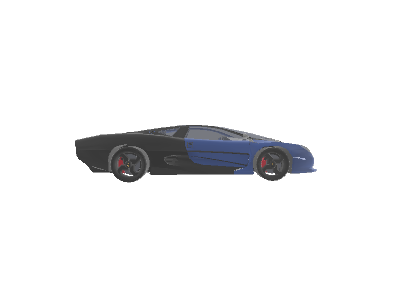}\vspace{-30pt}
\includegraphics[width=1\linewidth]{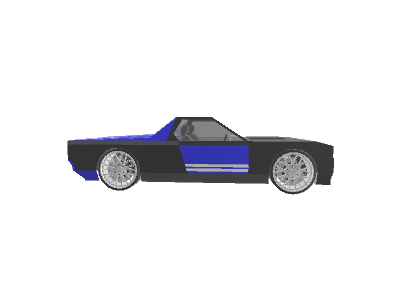}\vspace{-10pt}
\end{minipage}}
\subfloat[GT Points]{
\begin{minipage}[b]{0.15\linewidth}
\includegraphics[width=1\linewidth]{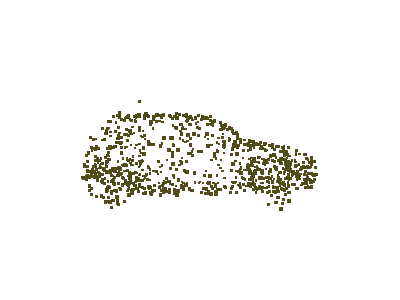}\vspace{-30pt}
\includegraphics[width=1\linewidth]{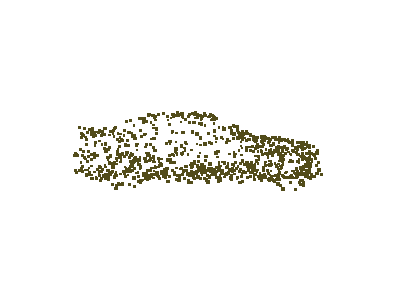}\vspace{-30pt}
\includegraphics[width=1\linewidth]{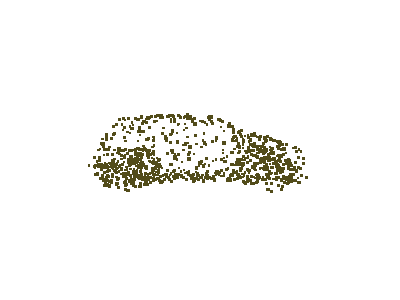}\vspace{-30pt}
\includegraphics[width=1\linewidth]{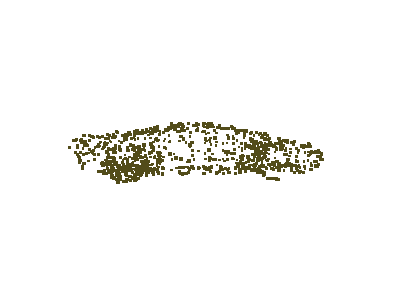}\vspace{-30pt}
\includegraphics[width=1\linewidth]{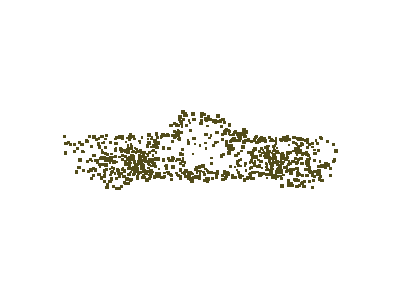}\vspace{-10pt}
\end{minipage}}
\subfloat[Our GN]{
\begin{minipage}[b]{0.15\linewidth}
\includegraphics[width=1\linewidth]{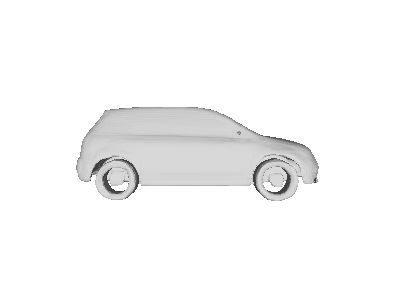}\vspace{-30pt}
\includegraphics[width=1\linewidth]{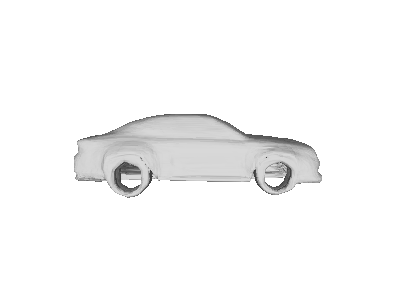}\vspace{-30pt}
\includegraphics[width=1\linewidth]{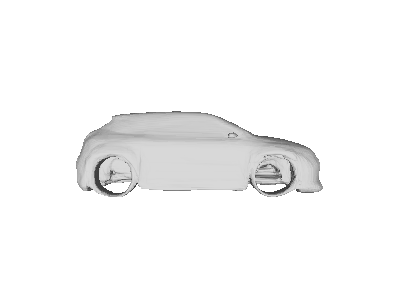}\vspace{-30pt}
\includegraphics[width=1\linewidth]{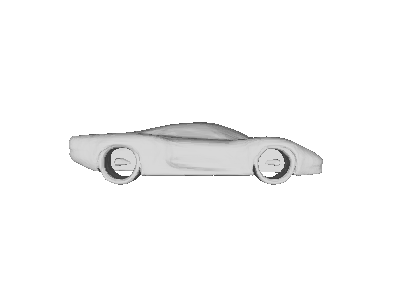}\vspace{-30pt}
\includegraphics[width=1\linewidth]{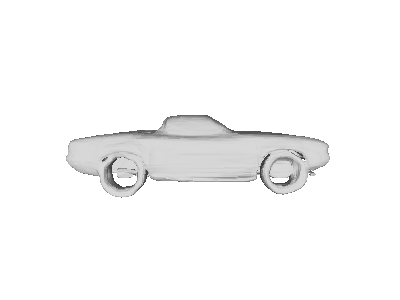}\vspace{-10pt}
\end{minipage}}
\subfloat[Our Adam]{
\begin{minipage}[b]{0.15\linewidth}
\includegraphics[width=1\linewidth]{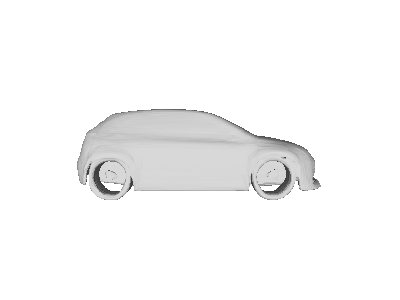}\vspace{-30pt}
\includegraphics[width=1\linewidth]{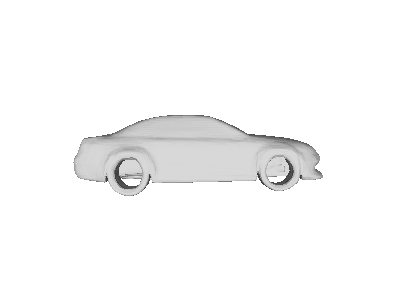}\vspace{-30pt}
\includegraphics[width=1\linewidth]{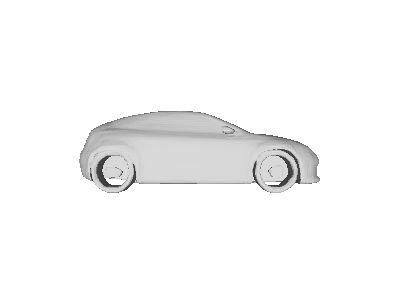}\vspace{-30pt}
\includegraphics[width=1\linewidth]{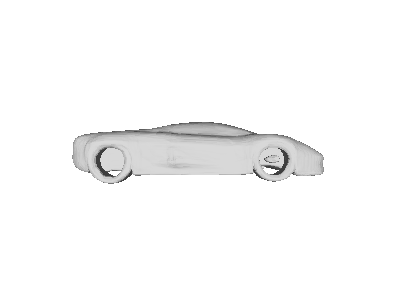}\vspace{-30pt}
\includegraphics[width=1\linewidth]{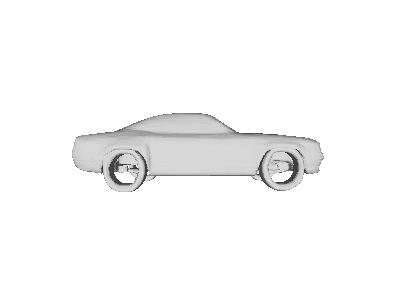}\vspace{-10pt}
\end{minipage}}
\subfloat[DeepSDF GN]{
\begin{minipage}[b]{0.15\linewidth}
\includegraphics[width=1\linewidth]{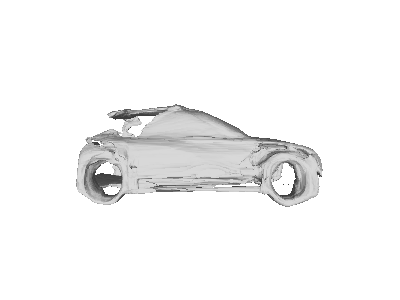}\vspace{-30pt}
\includegraphics[width=1\linewidth]{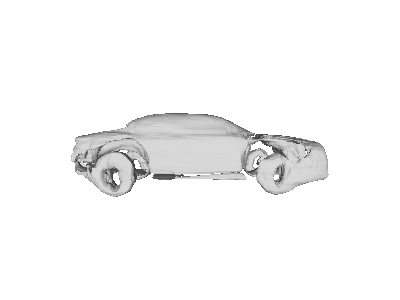}\vspace{-30pt}
\includegraphics[width=1\linewidth]{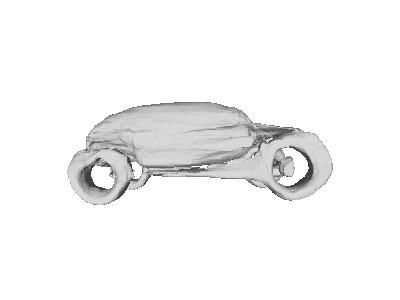}\vspace{-30pt}
\includegraphics[width=1\linewidth]{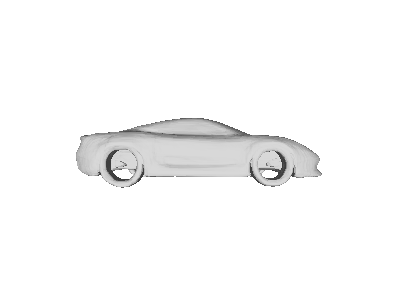}\vspace{-30pt}
\includegraphics[width=1\linewidth]{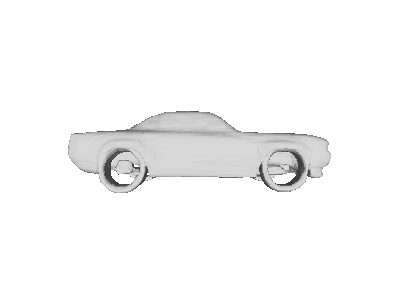}\vspace{-10pt}
\end{minipage}}
\subfloat[DeepSDF Adam]{
\begin{minipage}[b]{0.15\linewidth}
\includegraphics[width=1\linewidth]{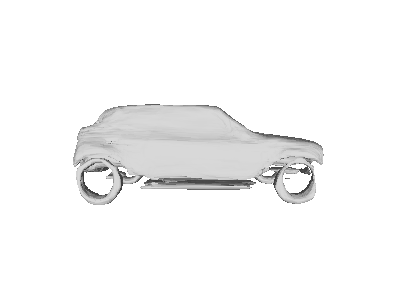}\vspace{-32pt}
\includegraphics[width=1\linewidth]{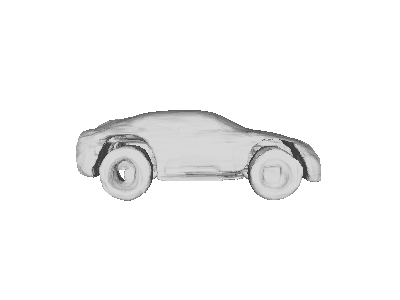}\vspace{-30pt}
\includegraphics[width=1\linewidth]{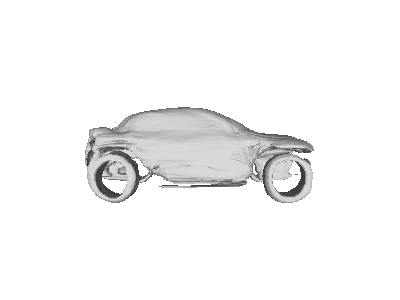}\vspace{-30pt}
\includegraphics[width=1\linewidth]{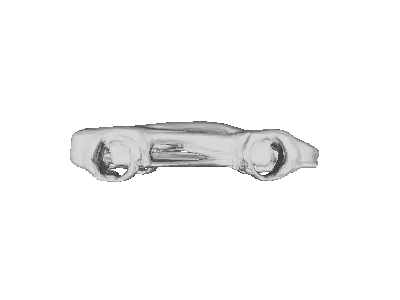}\vspace{-30pt}
\includegraphics[width=1\linewidth]{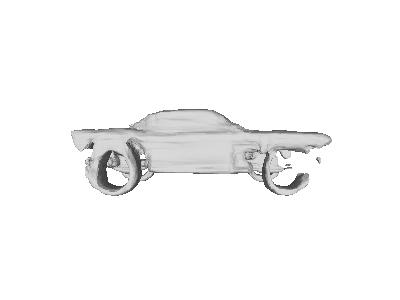}\vspace{-10pt}
\end{minipage}}

\caption{Shape optimization results for complete point clouds taken from ShapeNet. From left to right: Original model, point cloud samples, our proposed generator with normalizing flow optimized with Gauss-Newton, the same generator optimized with Adam, DeepSDF results using Gauss-Newton, and DeepSDF results using Adam optimizer.}
\label{fig: shapenet complete}
\vspace{-10pt}
\end{figure*}

\subsubsection{Test on ShapeNet with Complete Point Clouds}
For each analyzed object from ShapeNet, we extract surface points following the data preparation steps outlined in the original DeepSDF paper. 
We refer to this as the ``complete point cloud'', as it is an evenly distributed set of points lying on the surface of the object. The optimization simply employs the priorly mentioned surface distance evaluated for all points in the point cloud.

We use the Chamfer distance as an evaluation metric. We evaluate the bidirectional chamfer distance between the points sampled on the zero-level set of the optimized shape (set $S_1$) and the original surface points extracted from ShapeNet data (set $S_2$), i.e.
\begin{equation}
    \begin{aligned}
        d_{CD}(S_1, S_2) = \frac{1}{\vert S_1 \vert}\mathop{\sum}_{x \in S_1} \mathop{\min}_{y \in S_2} \Vert x-y \Vert_2^2 + \frac{1}{\vert S_2 \vert}\mathop{\sum}_{y \in S_2} \mathop{\min}_{x \in S_1} \Vert x-y \Vert_2^2
    \end{aligned}
    \label{bidirectional chamfer}
\end{equation}
We use 1000 surface points for shape reconstruction. Table~\ref{tab:shapenet complete} lists the results obtained for complete point clouds, note that for statistical purposes, all values have been multiplied by 1000. and Figure~\ref{fig: shapenet complete} shows qualitative examples. As can be observed, our method consistently produces reasonable shapes and delivers the best results in terms of the mean and standard deviation, thus indicating better robustness and convergence ability.
 
\begin{table}[t]
    \centering 
        \vspace{14pt}
    \caption{Quantitative results obtained on complete point clouds.}
    \begin{threeparttable} 
        \begin{tabular}{cccc} 
            \toprule       
            \multirow{2}{*}{\bf Method$\backslash$ Metric}
            &\multicolumn{3}{c}{\bf Chamfer Distance } \cr
            &{\bf Median} & {\bf Mean} & {\bf Std}  \cr
            \midrule  
            Our GN & 0.2023 & \textbf{0.2588} & \textbf{0.2785} \cr 
            Our Adam & 0.2078 & 0.2905  & 0.3231   \cr
            DeepSDF GN & \textbf{0.1803} & 0.2921 & 0.3339 \cr 
            DeepSDF & 0.2201 & 0.3072 & 0.3343 \cr 
            \bottomrule 
        \end{tabular}
    \end{threeparttable}
    \label{tab:shapenet complete} 
    \vspace{7pt}
\end{table}

\begin{figure*}[htb]
\centering
\subfloat[GT obj]{
\begin{minipage}[b]{0.15\linewidth}
\includegraphics[width=1\linewidth]{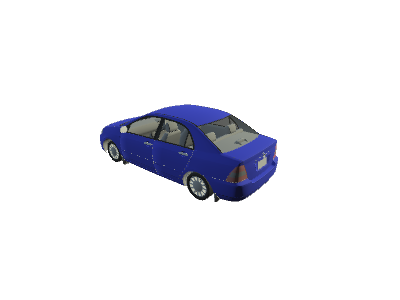}\vspace{-20pt}
\includegraphics[width=1\linewidth]{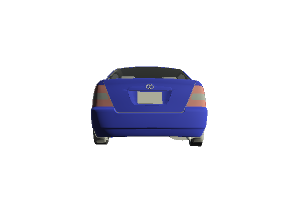}\vspace{-20pt}
\includegraphics[width=1\linewidth]{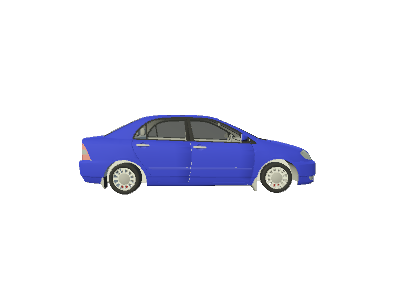}\vspace{-10pt}
\end{minipage}}
\subfloat[GT Points]{
\begin{minipage}[b]{0.15\linewidth}
\includegraphics[width=1\linewidth]{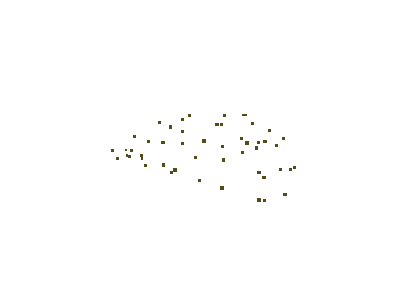}\vspace{-20pt}
\includegraphics[width=1\linewidth]{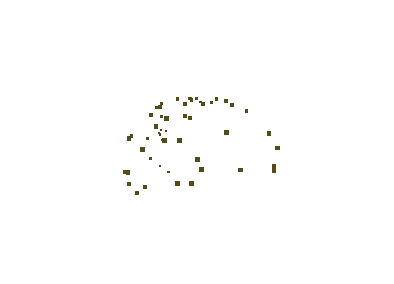}\vspace{-20pt}
\includegraphics[width=1\linewidth]{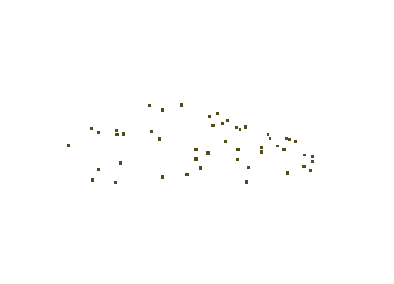}\vspace{-10pt}
\end{minipage}}
\subfloat[Our GN]{
\begin{minipage}[b]{0.15\linewidth}
\includegraphics[width=1\linewidth]{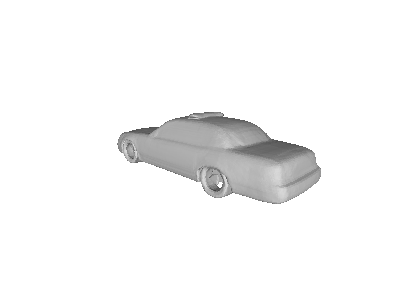}\vspace{-20pt}
\includegraphics[width=1\linewidth]{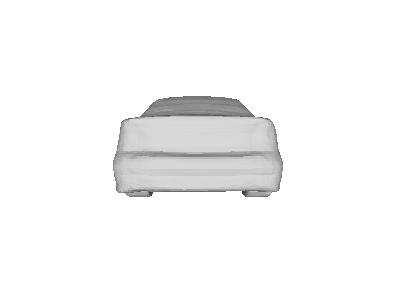}\vspace{-20pt}
\includegraphics[width=1\linewidth]{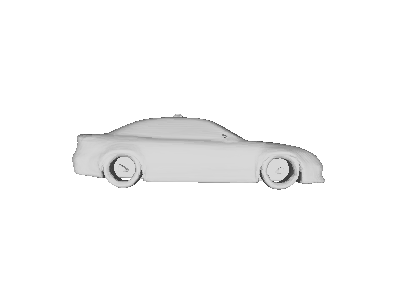}\vspace{-10pt}
\end{minipage}}
\subfloat[Our Adam]{
\begin{minipage}[b]{0.15\linewidth}
\includegraphics[width=1\linewidth]{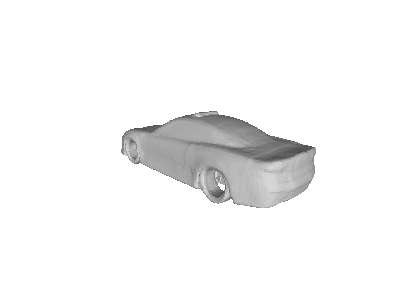}\vspace{-20pt}
\includegraphics[width=1\linewidth]{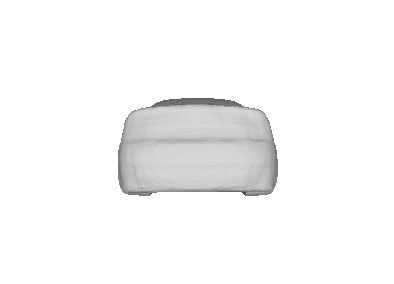}\vspace{-20pt}
\includegraphics[width=1\linewidth]{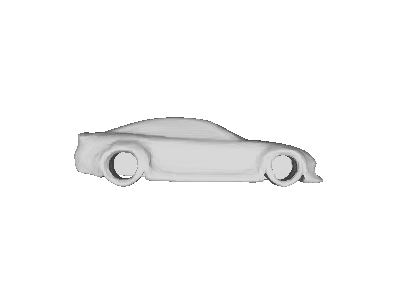}\vspace{-10pt}
\end{minipage}}
\subfloat[DeepSDF GN]{
\begin{minipage}[b]{0.15\linewidth}
\includegraphics[width=1\linewidth]{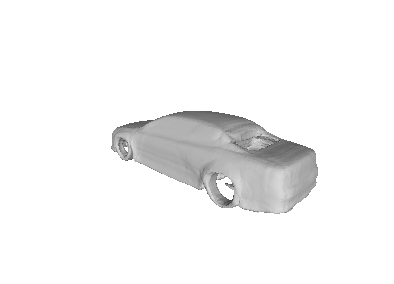}\vspace{-20pt}
\includegraphics[width=1\linewidth]{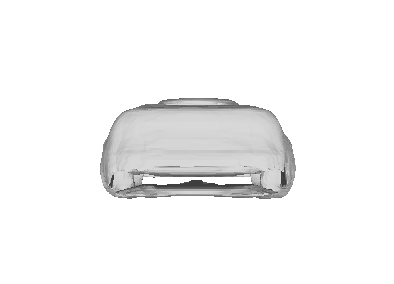}\vspace{-20pt}
\includegraphics[width=1\linewidth]{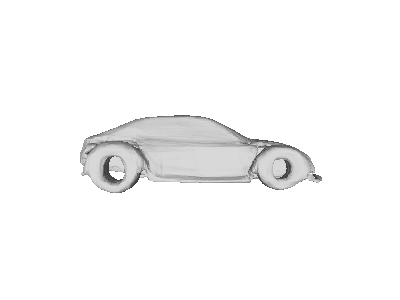}\vspace{-10pt}
\end{minipage}}
\subfloat[DeepSDF Adam]{
\begin{minipage}[b]{0.15\linewidth}
\includegraphics[width=1\linewidth]{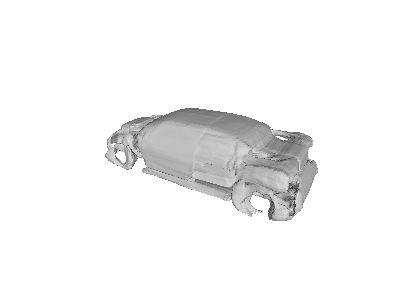}\vspace{-20pt}
\includegraphics[width=1\linewidth]{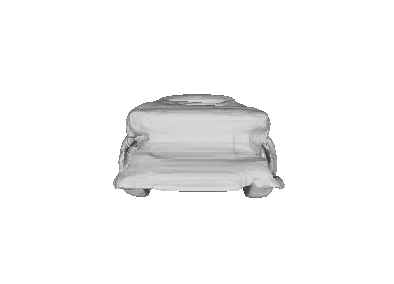}\vspace{-20pt}
\includegraphics[width=1\linewidth]{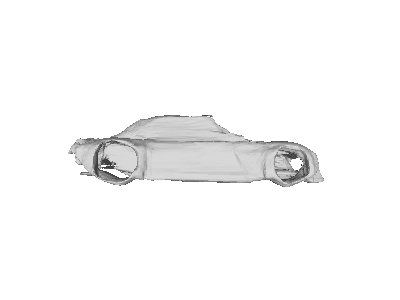}\vspace{-10pt}
\end{minipage}}
\caption{Shape optimization results for partial point clouds taken from ShapeNet. From left to right: Original model, point cloud samples, our proposed generator with normalizing flow optimized with Gauss-Newton, the same generator optimized with Adam, DeepSDF results using Gauss-Newton, and DeepSDF results using Adam optimizer.}
\label{fig: shapenet partial}
\end{figure*}

\subsubsection{Tests on ShapeNet with Partial Point Clouds}
\begin{table}[!bt]
    \centering 
    \caption{Quantitative results obtained on partial point clouds.}
    \begin{threeparttable} 
        \begin{tabular}{cccc} 
            \toprule       
            \multirow{2}{*}{\bf Method$\backslash$ Metric}
            &\multicolumn{3}{c}{\bf Chamfer Distance } \cr
            &{\bf Median} & {\bf Mean} & {\bf Std}  \cr
            \midrule 
            Our GN & 0.4017 & \textbf{0.4470} & 0.2017 \cr 
            Our Adam & 0.4128 & 0.4693 & \textbf{0.1880} \cr 
            DeepSDF GN & \textbf{0.3982} & 0.4760 & 0.2301 \cr 
            DeepSDF & 0.4809 & 0.5169 & 0.2107 \cr 
            \bottomrule 
        \end{tabular}
    \end{threeparttable}
    \label{tab:shapenet partial} 
    \vspace{-15pt}
\end{table}
In order to simulate real-world-like scenarios where only a portion of the point cloud is visible, we render partial point clouds from 10 different views and repeat the surface distance based optimization for all networks and optimization variants over these sample sets individually.
Even though we only use partial point clouds to perform the shape optimization, we continue to evaluate the same bi-directional chamfer distance to evaluate the complete generated shape. For each object, we notably compute the average chamfer distance across these 10 perspectives and use it as the final result for that object. Table \ref{tab:shapenet partial} again lists the median, mean, and standard deviation for all objects. As can be observed, we continue to exhibit low median errors and strong robustness as indicated by significantly lower mean and standard deviations. As further supported by the qualitative results in Figure \ref{fig: shapenet partial}, the addition of normalizing flow provokes the consistent generation of more reasonable shapes, in the case of partial point cloud samples.
Note that, in this experiment, only 50 surface points are used for reconstruction.

\subsection{Test on real world dataset}

To evaluate the full system, we do experiments on the KITTI odometry dataset. As our focus lies on the quality of the vehicle reconstruction, we assess results from shape completeness and accuracy. Shape completeness analyses whether or not the reconstructed shape represents a complete, reasonable vehicle rather than just wheels or a partitioned mesh with artefacts. The accuracy measures the proximity of the reconstructed part and the original point cloud. We test 10 sequences which contain a significant number of vehicles. We compare four methods: DSP-SLAM with lidar as input, DSP-SLAM* with only stereo images, NF-SLAM, and NF-SLAM* with only mask constraint. Note that in DSP-SLAM, a 64-dimensional latent vector is utilized as the shape representation, whereas in our approach, we employ a 16-dimensional latent vector.

\subsubsection{Completeness Evaluation}

In order to assess the completeness of the object shapes, we use the 3D IoU (Intersection over Union) which can provide a quantitative metric to assess the size and thus completeness of the reconstructed object shapes. For each reconstructed vehicle, we estimated its 3D bounding box and calculate the 3D IoU  with the 3D bounding box detected from the lidar point cloud.
%
%
For each sequence, we computed the median, mean, and standard deviation. Quantitative results for all 10 sequences are summarized in Table~\ref{tab:iou}.

\begin{figure*}[htbp]
\centering
\subfloat[KITTI images]{
\begin{minipage}[b]{0.45\linewidth}
\includegraphics[width=1\linewidth]{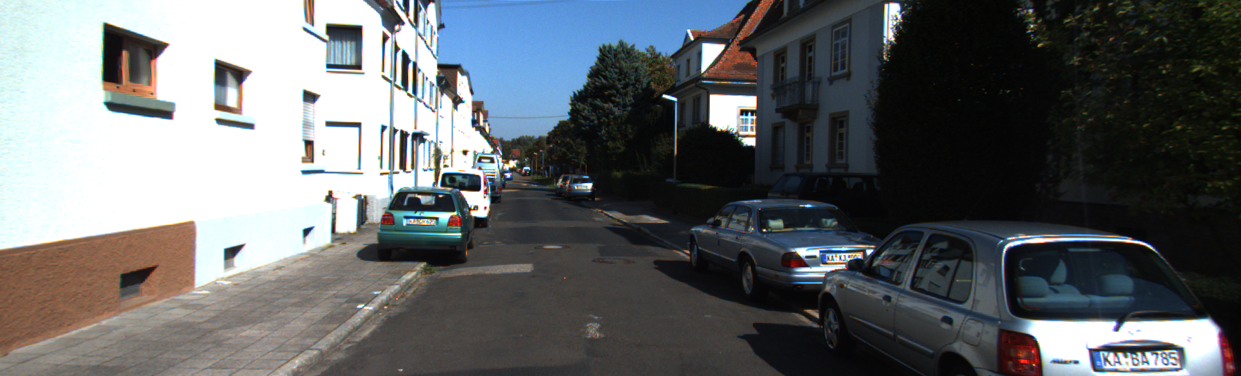}\vspace{10pt}
\includegraphics[width=1\linewidth]{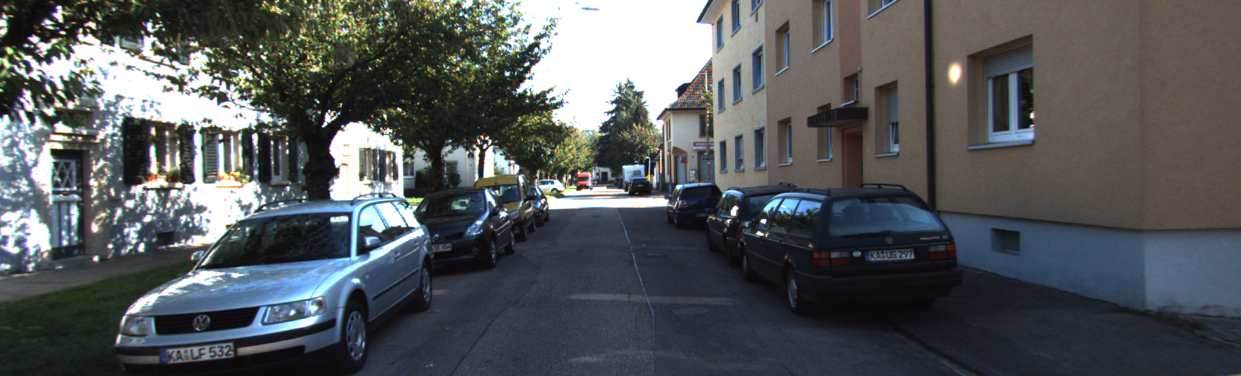}
\end{minipage}}
\subfloat[DSP-SLAM / DSP-SLAM*]{
\begin{minipage}[b]{0.25\linewidth}
\includegraphics[width=1\linewidth]{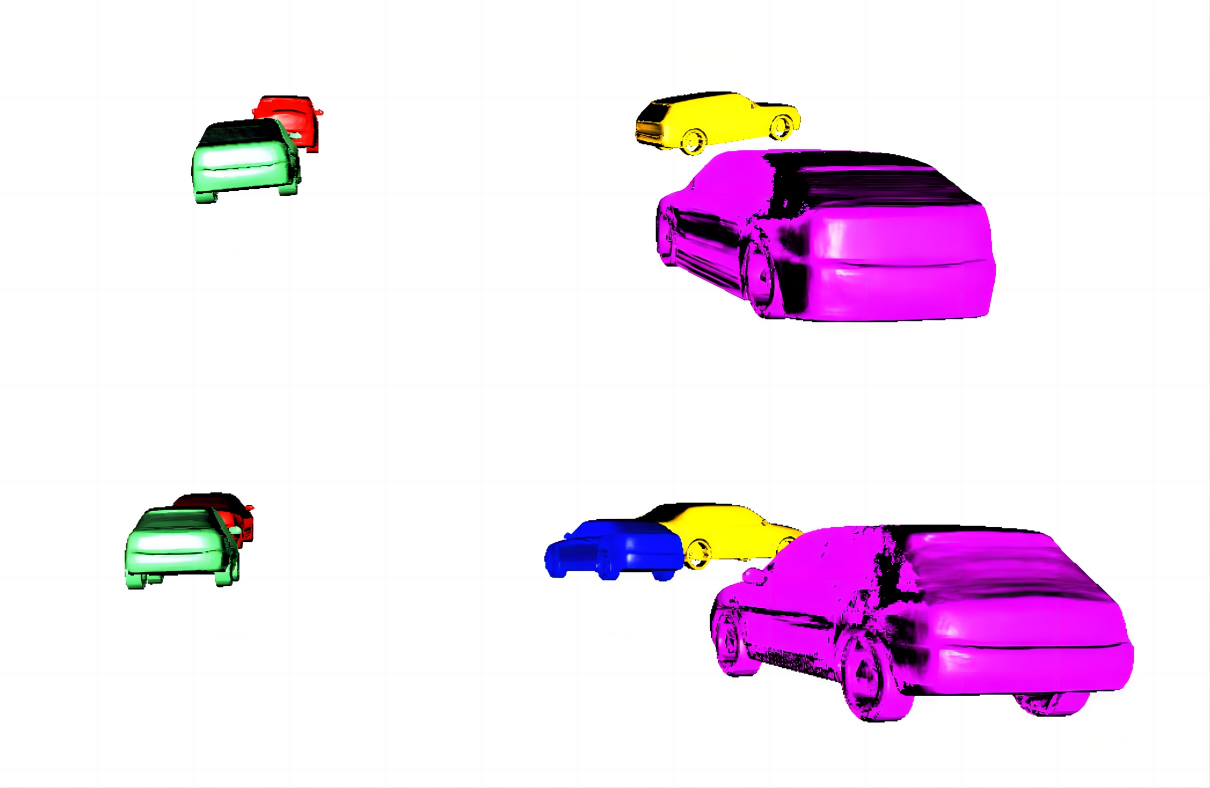}
\includegraphics[width=1\linewidth]{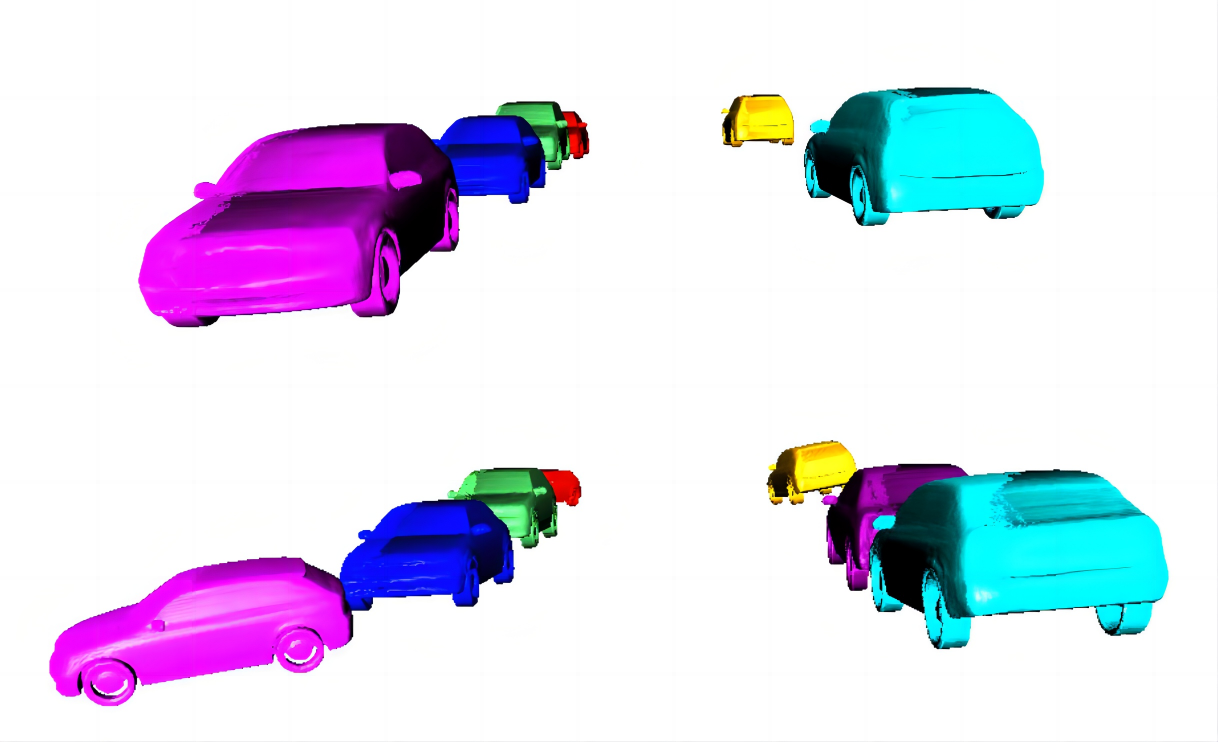}
\end{minipage}}
\subfloat[NF-SLAM / NF-SLAM*]{
\begin{minipage}[b]{0.25\linewidth}
\includegraphics[width=1\linewidth]{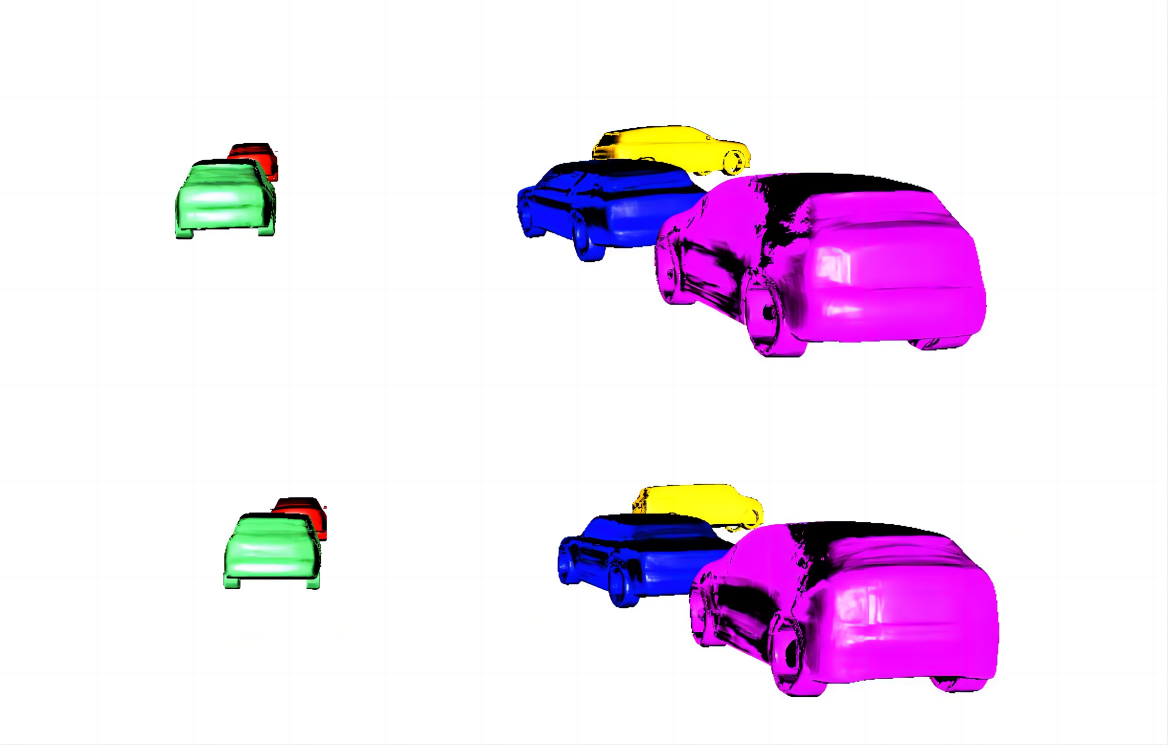}
\includegraphics[width=1\linewidth]{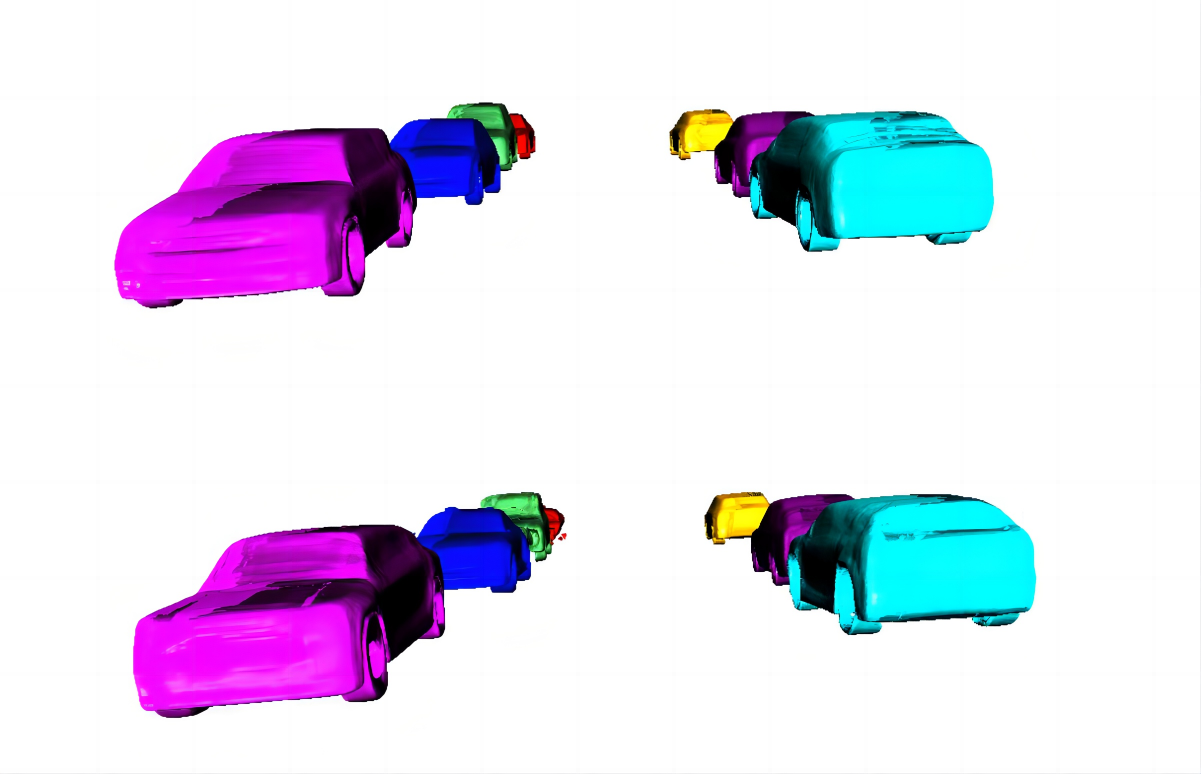}
\end{minipage}}
\caption{Qualitative results obtained on KITTI dataset. In subfigure (b), DSP-SLAM is positioned above, DSP-SLAM* is positioned below. In subfigure (c), NF-SLAM is positioned above, NF-SLAM* is positioned below.}
\label{fig:kitti}
\vspace{14pt}
\end{figure*}

\begin{table*}[!htb]
    \centering 
    \caption{Assessment of object completeness using the 3D IoU between the inferred and ground truth bounding boxes.}
    \begin{threeparttable} 
\begin{tabular}{lrrrrrrrrrrrr}
 \toprule 
{} & \multicolumn{3}{c}{\bf DSP-SLAM} & \multicolumn{3}{c}{\bf DSP-SLAM*} & \multicolumn{3}{c}{\bf NF-SLAM} & \multicolumn{3}{c}{\bf NF-SLAM*} \\

{\bf Seq} & {\bf Median} & {\bf Mean} & {\bf Std} & {\bf Median} & \bf {Mean} & {\bf Std} & {\bf Median} & {\bf Mean} & {\bf Std} & {\bf Median} & {\bf Mean} & {\bf Std} \\
\midrule

00 & 0.8059 & 0.8004 & \bf{0.0528} & 0.7134 & 0.6265 & 0.2064 & \bf{0.8402} & \bf{0.8284} & 0.0733 & 0.8092 & 0.7810 & 0.1257 \\
05 & 0.7855 & 0.7855 & 0.0619 & 0.7656 & 0.7344 & 0.1559 & \bf{0.8426} & \bf{0.8350} & \bf{0.0528} & 0.8034 & 0.7925 & 0.0992 \\
06 & 0.7943 & \bf{0.7960} & \bf{0.0525} & 0.7682 & 0.7161 & 0.1391 & \bf{0.8078} & 0.7816 & 0.1140 & 0.7876 & 0.7763 & 0.0989 \\
07 & 0.8058 & 0.7902 & 0.0845 & 0.7341 & 0.6670 & 0.1992 & \bf{0.8115} & \bf{0.8087} & \bf{0.0647} & 0.8070 & 0.7855 & 0.1150 \\
08 & 0.8142 & 0.8050 & 0.0596 & 0.7667 & 0.6525 & 0.2572 & \bf{0.8394} & \bf{0.8326} & \bf{0.0574} & 0.8228 & 0.8040 & 0.1099 \\
10 & 0.8018 & \bf{0.7931} & \bf{0.0659} & 0.7586 & 0.6542 & 0.2295 & \bf{0.8241} & 0.7888 & 0.1373 & 0.8037 & 0.7436 & 0.1749 \\
11 & 0.7817 & 0.7813 & \bf{0.0416} & 0.7768 & 0.7006 & 0.2071 & \bf{0.8440} & \bf{0.8425} & 0.0607 & 0.7966 & 0.7943 & 0.0829 \\
15 & 0.8097 & 0.8039 & 0.0570 & 0.7349 & 0.6526 & 0.2220 & 0.8189 & \bf{0.8316} & \bf{0.0512} & \bf{0.8325} & 0.8038 & 0.1168 \\
16 & 0.8090 & 0.8024 & \bf{0.0477} & 0.7802 & 0.7267 & 0.1427 & \bf{0.8471} & \bf{0.8289} & 0.0812 & 0.8060 & 0.7803 & 0.1372 \\
19 & 0.8080 & 0.7965 & 0.0818 & 0.7444 & 0.6775 & 0.1799 & \bf{0.8319} & \bf{0.8289} & \bf{0.0617} & 0.8099 & 0.7884 & 0.1078 \\

\bf{Mean} & 0.8015 &  0.7954 & \bf{0.0605} & 0.7543 & 0.6808 &  0.1939 & \bf{0.8307} & \bf{0.8207} &   0.0754 & 0.8078 & 0.7849 & 0.1168 \\
            \bottomrule 
        \end{tabular}
    \end{threeparttable}
    \label{tab:iou}
    \vspace{10pt}
\end{table*}

\begin{table*}[!hbt]
    \centering 
    \caption{Unidirectional Chamfer Distances evaluating the accuracy of the reconstructed shapes on KITTI.}
    \begin{threeparttable} 
\begin{tabular}{lrrrrrrrrrrrr}
 \toprule 
{} & \multicolumn{3}{c}{\bf DSP-SLAM} & \multicolumn{3}{c}{\bf DSP-SLAM*} & \multicolumn{3}{c}{\bf NF-SLAM} & \multicolumn{3}{c}{\bf NF-SLAM*} \\

{\bf Seq} & {\bf Median} & {\bf Mean} & {\bf Std} & {\bf Median} & \bf {Mean} & {\bf Std} & {\bf Median} & {\bf Mean} & {\bf Std} & {\bf Median} & {\bf Mean} & {\bf Std} \\
\midrule

00 & 0.2517 & 0.3266 & \bf{0.4428} & 1.1405 & 3.1542 & 5.3138 & \bf{0.2171} & \bf{0.3218} & 0.5123 & 0.2657 & 0.4961 & 0.8832 \\
05 & \bf{0.2658} & 0.3449 & 0.2268 & 0.3386 & 0.7271 & 1.0228 & 0.3013 & \bf{0.3183} & \bf{0.1700} & 0.3303 & 0.5548 & 0.8193 \\
06 & \bf{0.2225} & \bf{0.2660} & \bf{0.1652} & 0.5065 & 2.3795 & 8.5699 & 0.2348 & 1.1300 & 5.9138 & 0.3089 & 0.4560 & 0.5099 \\
07 & \bf{0.2779} & \bf{0.3243} & \bf{0.2192} & 0.6718 & 2.0225 & 5.8163 & 0.2985 & 0.3838 & 0.3329 & 0.3620 & 0.4558 & 0.4943 \\
08 & 0.3040 & 0.3008 & \bf{0.1057} & 0.4501 & 1.5392 & 2.1141 & \bf{0.2235} & \bf{0.2698} & 0.1619 & 0.2871 & 0.3831 & 0.3104 \\
10 & 0.3365 & \bf{0.4190} & \bf{0.3355} & 0.6131 & 2.9308 & 7.9568 & \bf{0.2211} & 0.7863 & 1.9294 & 0.3111 & 1.7130 & 4.8879 \\
11 & \bf{0.2479} & \bf{0.3066} & 0.1751 & 0.4041 & 1.0182 & 1.5594 & 0.3009 & 0.3085 & \bf{0.1650} & 0.3174 & 0.3391 & 0.1792 \\
15 & 0.3049 & 0.3855 & 0.3959 & 0.4148 & 1.4261 & 2.1337 & \bf{0.2628} & \bf{0.2819} & \bf{0.1121} & 0.2802 & 0.9101 & 3.7743 \\
16 & 0.2333 & \bf{0.2688} & 0.2224 & 0.3421 & 1.0811 & 1.9455 & \bf{0.2329} & 0.2781 & \bf{0.1673} & 0.2575 & 0.4779 & 0.5647 \\
19 & 0.2341 & 0.2549 & \bf{0.1349} & 0.5741 & 2.3888 & 5.5756 & \bf{0.2158} & \bf{0.2467} & 0.1522 & 0.2494 & 0.3449 & 0.3482 \\

\bf{Mean} & 0.2678 &  \bf{0.3197} &  \bf{0.2423} & 0.5455 & 1.8667 & 4.2007 & \bf{0.2508} &  0.4325 & 0.9616 &  0.2969 &  0.6130 & 1.2771 \\

            \bottomrule 

        \end{tabular}
    \end{threeparttable}
\label{tab:uni chamfer}
    \vspace{-10pt}
\end{table*}

\subsubsection{Accuracy Evaluation}
To conclude, the lidar point clouds--- captured by a Velodyne laser scanner and utilized to create the ground-truth data in the KITTI dataset---can be used for a more detailed assessment of the accuracy of the reconstructed part of the shape for each vehicle. 
To accomplish this, we transform the lidar point cloud into individual object coordinate frames and then segment out the subset of the points that resides within the vehicle's 3D reference bounding box.

Owing to the nature of real-world scenarios, only a partial view of the car is visible at any given moment. This implies that our observations are limited to partial point samples. Hence, we opted for the unidirectional chamfer distance as our evaluation criterion and find the distance for each of our input samples to the ground truth vehicle shape, i.e.
\begin{equation}
    \begin{aligned}
        d_{UCD}(S_1, S_2) = \frac{1}{\vert S_1 \vert}\mathop{\sum}_{x \in S_1} \mathop{\min}_{y \in S_2} \Vert x-y \Vert_2^2 
    \end{aligned}
    \label{unidirectional chamfer}
\end{equation}

Quantitative results are summarized in Table~\ref{tab:uni chamfer}. For the sake of convenience, the numerical values in the table have been multiplied by 100. Qualitative results are shown in Figure~\ref{fig:kitti}. From the experimental results, it can be observed that when the input is limited to only stereo images, NF-SLAM is comparable to DSP-SLAM using lidar data in most cases. NF-SLAM* achieves decent results using only mask loss, whereas the performance of the DSP-SLAM* significantly decreases.

\section{CONCLUSION}

We propose a novel implicit shape generation model making use of normalizing flow for improved robustness, convergence and overall performance in situations of limited input samples. Combined with a multi-frame optimization scheme, the module achieves reasonable vehicle shape reconstructions on par with lidar-based optimization schemes. We believe the proposed architecture to be of potentially broad interest in vision-only object-level perception tasks, both with and without a focus on automotive applications.

\section*{ACKNOWLEDGMENT}
We would like to acknowledge the funding support provided by project 62250610225 by the Natural Science Foundation of China, as well as projects 22DZ1201900, 22ZR1441300, and dfycbj-1 by the Natural Science Foundation of Shanghai.






\bibliographystyle{IEEEtran}
\bibliography{main} 
\end{document}